\documentclass[runningheads]{llncs}
\usepackage{graphicx}
\usepackage{amsmath,amssymb} 
\usepackage{color}
\usepackage{epstopdf}

\begin{document}
\pagestyle{headings}
\mainmatter

\def\ACCV18SubNumber{777}  
\def\eg{\emph{e.g.}}
\def\Eg{\emph{E.g.}}
\def\etal{\emph{et al.}}

\title{SingleGAN: Image-to-Image Translation by a Single-Generator Network  using Multiple Generative Adversarial Learning} 

\titlerunning{SingleGAN}
\authorrunning{Xiaoming~Yu, Xing~Cai, Zhenqiang~Ying, Thomas~Li, and Ge~Li}

\newcommand*\samethanks[1][\value{footnote}]{\footnotemark[#1]}

\author{
	Xiaoming~Yu\thanks{indicates equal contribution},
	Xing~Cai\samethanks[1],
	Zhenqiang~Ying,
	Thomas~Li,
	\and Ge~Li
}
\institute{School of Electronic and Computer Engineering, Shenzhen Graduate School, Peking University, Shenzhen, China}

\maketitle

\begin{abstract}
Image translation is a burgeoning field in computer vision where the goal is to learn the mapping between an input image and an output image. 
However, most recent methods require multiple generators for modeling different domain mappings, which are inefficient and ineffective on some multi-domain image translation tasks.
In this paper, we propose a novel method, SingleGAN, to perform multi-domain image-to-image translations with a single generator.
We introduce the domain code to explicitly control the different generative tasks and integrate multiple optimization goals to ensure the translation.
Experimental results on several unpaired datasets show superior performance of our model in translation between two domains. 
Besides, we explore variants of SingleGAN for different tasks, including one-to-many domain translation, many-to-many domain translation and one-to-one domain translation with multimodality.
The extended experiments show the universality and extensibility of our model.
Code is available at~\url{https://github.com/Xiaoming-Yu/SingleGAN}.
\end{abstract}

\section{Introduction}
\label{sec:intro}
Recently, more and more attention has been paid to image-to-image translation due to its exciting potential in a variety of image processing applications~\cite{isola2017pix2pix}.
Although existing methods show impressive results on one-to-one mapping problems,
they need to build multiple generators for modeling multiple mappings,
which are inefficient and ineffective in some multi-domain and multi-model image translation tasks.
Intuitively, many multi-mapping translation tasks are not independent and share some common features such as scene contents in transformations between different seasons.
By sharing a network between related tasks, we can enable our model to generalize better on each separated task. 
In this paper, we propose a single-generator generative adversarial network (GAN), called SingleGAN, 
to solve multi-mapping translation tasks effectively and efficiently.
To indicate a specific mapping, we introduce the domain code as an auxiliary input to the network.
Then we integrate multiple optimization goals to learn each specific translation.

As illustrated in Fig.~\ref{fig:base}, the base SingleGAN model is utilized to learn the bijection between two domains.
Since each domain dataset is not required to have the label of other domains,
SingleGAN can make full use of the existing different datasets to learn the multi-domain translation. 

To explore the potential and generality of SingleGAN, we also extend it to three cross domains translation tasks, which are more complex and practical. 
The first variant model tries to address the one-to-many domain translation task that processes a source domain input to a different target domains,
such as the image style transfer.
The Second model explore the many-to-many domain translation task.
Unlike the recent method~\cite{choi2017stargan} requires detailed annotation of category information to training the auxiliary classifier,
we use multiple adversarial objects to help network captures different domain distribution separately.
It means that SingleGAN is capable of learning multi-domain mappings by weakly supervised learning since we do not need to label all the training data with detailed annotation.
The third variant model attempts to increase the generative diversity by introducing attribute latent code.
A similar idea is used in BicycleGAN~\cite{zhu2017multimodal} to address the multimodal translation problem.
Our third model can be considered a generalization of BicycleGAN towards unpaired image-to-image translation.

To summarize, our contributions are as follows: 
\begin{itemize}
	\item We propose SingleGAN, a novel GAN that utilizes a single generator and a group of discriminators to accomplish the unpaired image-to-image translation.
	\item We show the generality and flexibility of SingleGAN by extending it to achieve three different kinds of translation tasks.
	\item Experimental results demonstrate that our approach is more effective and general-purpose than several state-of-art methods.
\end{itemize}

\section{Related work}
\label{sec:related}

\subsection{Generative Adversarial Networks}
Influenced by a zero-sum game,  
a typical GAN model consists of two modules: a generator and a discriminator. 
While the discriminator learns to distinguish between real and fake samples, the generator learns to generate fake samples that are indistinguishable from real samples.
The GANs have shown impressive results in various computer vision tasks such as image generation, image editing~\cite{brock2016neural} and representation learning~\cite{radford2015dcgan}. 
Recently, GAN-based conditional image generation has also been actively studied. 
To be specific, the various of extension GANs have achieved good results in many generation tasks such as image inpainting~\cite{pathak2016context}, super-resolution~\cite{ledig2016photoresolution}, text2image\cite{reed2016text2image}, as well as to other domains such as videos~\cite{Vondrick2016Generating} and 3D data\cite{Wu2016Learning}. 
In this paper, we propose a scalable GAN framework to achieve image translation based on conditional image generation. 

\subsection{Image-to-Image Translation}
The idea of image-to-image translation goes back to Image Analogies~\cite{Hertzmann2001Image}, in which Hertzmann \etal ~proposed a network to transfer the texture information from a source modality space onto a target modality space.
Image-to-image translation has received more attention since the flourishing growth of GANs. 
The pioneering work, Pix2pix~\cite{isola2017pix2pix} uses cGAN\cite{mirza2014conditional} to perform supervised image translation from paired data. 
As those methods adopt supervised learning, 
sufficient paired data are required to train the network.
However,
preparing paired images can be time-consuming and laborious
(\eg ~artistic stylization) and even impossible for some applications (\eg ~male to female face transfiguration).
To address this issue,
For example, CycleGAN~\cite{CycleGAN2017}, DiscoGAN~\cite{kim2017disco} and DualGAN~\cite{Yi2017DualGAN} introduce a cycle-consistency constraint, which widely used in visual tracking~\cite{Sundaram2010Dense} and language domain~\cite{Brislin1970Back}, to learn convincing mappings across image domains from unpaired images.
Based on a shared-latent space assumption,
UNIT~\cite{liu2017UNIT} extends the Coupled GAN~\cite{liu2016coupled} to learn a joint distribution of different domains without paired images. 
FaderNet~\cite{lample2017fader} is also successful in the controlling of attributes by adding the discriminator to the latent space.  Even though, these methods have promoted the development of one-to-one mapping image translation, they have limitations in scalability for multi-mapping translation. 
By introducing an auxiliary classifier in the discriminator,
StarGAN~\cite{choi2017stargan} achieved translation among different facial attributes with a single generator.
However, this method may learn an inefficient domain mapping when the attribute labels are not sufficient for training the auxiliary classifier even if it introduces a mask vector.

\section{Base Model}
\label{sec:method}
\begin{figure*}[b]
\centering
\includegraphics[width=.95\linewidth]{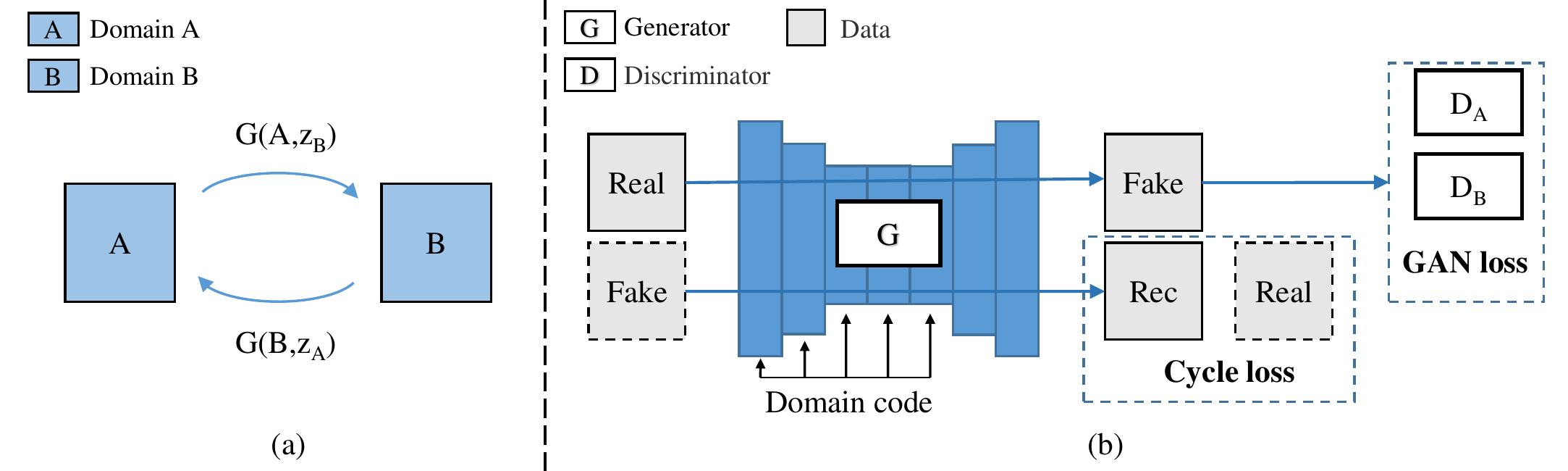}
\caption{(a) The base model contains two mapping direction: $\text{A} \rightarrow \text{B}$ and $\text{B} \rightarrow \text{A}$.
	(b) Our base model architecture, which consists of a generator and a group of discriminators. 
}
\label{fig:base}
\end{figure*}

The main architecture is shown in Fig.~\ref{fig:base}. In order to take advantage of the correlation between two related tasks, SingleGAN adopts a single generator to achieve bi-direction translation. 

The goal of the model is to learn a mapping $G :\text{A} \leftrightarrow \text{B}$. By adding the domain code, $G$ is redefined as
\begin{align}
\begin{split}
x_{\text{B}}^{fake} = G (x_{\text{A}}, z_{\text{B}}), \\
x_{\text{A}}^{fake} = G (x_{\text{B}}, z_{\text{A}}),
\end{split}
\end{align}
where $x^{fake}_{{*}}$ is the fake sample generated by the generator, sample $x_{{*}}$  belongs to the set of domain $\chi_{{*}}$ and $z_{\text{A}}, z_{\text{B}}$ are domain code for domain \text{A} and domain \text{B} respectively. 

\subsection{Domain Code Injection}
\label{sec:domain_code}
For capturing the distribution of different domains with a single generator,
it is necessary to indicate the mapping with auxiliary information.
Therefore, we introduce the domain code  $z$ to label the different mapping in the generator.
The domain code is constructed as a one-hot vector and similar to the latent code that is widely used to indicated the attributes of generated image~\cite{chen2016infogan,zhu2017multimodal}.

Recent work~\cite{yuEccv2018} shows that different injection methods of latent code will effect the performance of generation model.
So we adopt the central biasing instance normalization (CBIN) proposed in~\cite{yuEccv2018} to inject the domain code in our SingleGAN model.
CBIN is defined as
\begin{equation}
	\text{CBIN} ( x_i) = \frac{x_i - \text{E}[x_i] }{ \sqrt{ \text{Var}[{x}_i] }} + \tanh(f_i(z)),
\end{equation}
where $i$ is the index of feature maps, $f_i$ is affine transformation applied on the domain code $z$ and its parameters are learned for each feature map $x_i$ in one layer. 
The CBIN aims to adjust the different distributions of input feature maps adaptively with learnable parameters, which makes the domain code able to manage the different tasks. 
Meanwhile, the distance between the different distributions of input data is also trainable, which means that the coupling degree of different tasks is determined by the model itself. 
This advantage enables different tasks to share parameters better, so as to promote each other better.

\subsection{Loss Functions}

\subsubsection{GAN Loss.}
Since our single generator has multi-domain outputs, we set up adversarial objectives for each target domain and employ a group of discriminators. The corresponding discriminator is used to identify the generated images in one domain. The adversarial loss is defined as
\begin{equation}
	\begin{split}
	\mathcal{L}_{adv}(G,D_{ \text{A}  }) = &\mathbb{E}_{\chi_{\text{A}}} [\log(D_{\text{A}}(x_{\text{A}}))] + \mathbb{E}_{\chi_{\text{B}}}[\log(1-D_{\text{A}}(G(x_{\text{B}},z_{\text{A}})))], \\
	\mathcal{L}_{adv}(G,D_{ \text{B}  }) = &\mathbb{E}_{\chi_{\text{B}}} [\log(D_{\text{B}}(x_{\text{B}}))] + \mathbb{E}_{\chi_{\text{A}}}[\log(1-D_{\text{B}}(G(x_{\text{A}},z_{\text{B}})))].
	\end{split}
\end{equation}

By optimizing multiple generative adversarial objectives,
the generator recovers different domain distributions that indicated by domain code $z$.

\subsubsection{Cycle Consistency Loss.}
Although the above GAN loss can complete domain translation, highly under-constrained mapping often leads to a mode collapse.
There are many possible mappings that can be inferred without the use of pairing information.

To reduce the space of the possible mappings, we use the cycle-consistency constraint~\cite{CycleGAN2017,choi2017stargan} in the training stage.
The cycle consistency loss is defined as $ x_{\text{A}} \approx G(G(x_{\text{A}},z_{\text{B}}), z_{\text{A}})$ and $ x_{\text{B}} \approx G(G(x_{\text{B}},z_{\text{A}}), z_{\text{B}})$ in our model. 
The formula of the cycle consistency loss is defined as  
\begin{equation}
\begin{split}
\mathcal{L}_{cyc}(G) =&  
\mathbb{E}_{\chi_{\text{A}} } \big[  \left \| x_{\text{A}} -  G(G(x_{\text{A}},z_{\text{B}}), z_{\text{A}})  \right \| _{1}  \big] + \\ 
&\mathbb{E}_{\chi_{\text{B}} } \big[  \left \| x_{\text{B}} - G(G(x_{\text{B}},z_{\text{A}}), z_{\text{B}})  \right \| _{1}  \big],
\end{split}
\end{equation}
where $  \left \| \cdot \right \| _{1}$ denotes $\ell_1$ norm.

\subsubsection{Full Objective.}
The final objective function is defined as 
\begin{equation}
	G^* = arg \min_G \max_{D_A, D_B} \sum_{i\in \{ \text{A}, \text{B}\} } \mathcal{L}_{adv}(G,D_i) +\lambda_{cyc} \cdot \mathcal{L}_{cyc}(G)
\end{equation}
where $\lambda_{cyc}$ controls the relative importance of the two objectives. 

\section{Extended models}

\begin{figure*}[b]
\centering
\includegraphics[width=.95\linewidth]{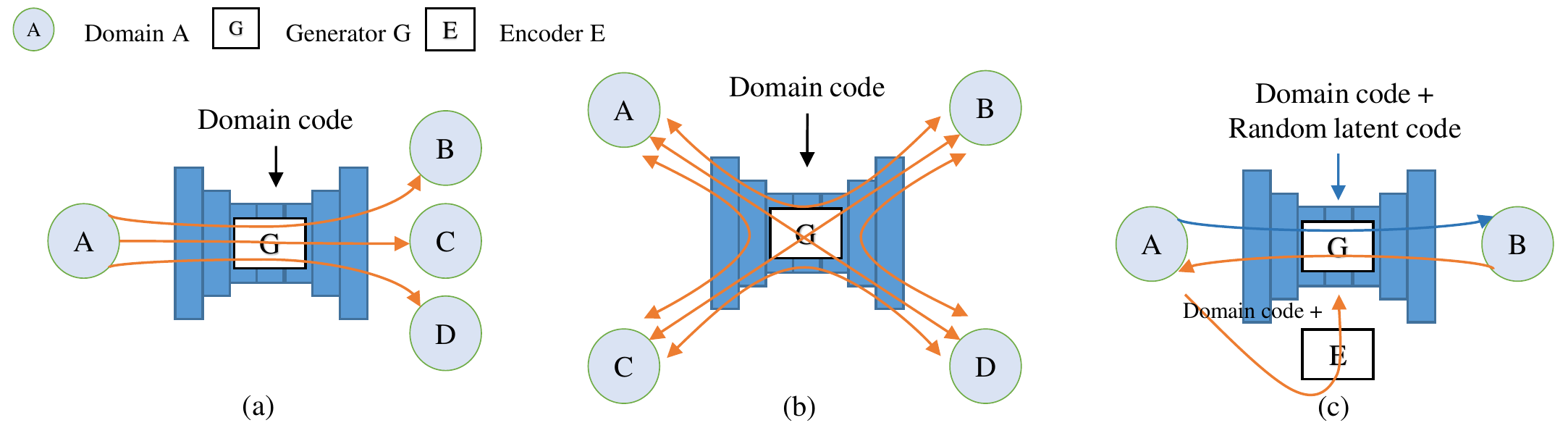}
\caption{Extended models: (a) one-to-many domain translation, (b) many-to-many domain translation, (c) one-to-one domain translation with multi-modal mapping.
}
\label{fig:extent}
\end{figure*}

To explore the potential and generality of SingleGAN, based on the above model, we extend three variants of our model to different tasks: one-to-many domain translation, many-to-many domain translation and one-to-one domain translation with multi-modal mapping.

\subsection{One-to-Many Domain Translation}
The first trial in Fig.\ref{fig:extent}(a) applies to unidirectional tasks, for example multi-task  detection and image multi-style transfer. 
As far as image style transfer is concerned, 
different style transfer from a single input image is a representative task of sharing semantics.
Our model shares the same texture information of the input image and apply different styles on it. 
Compared with traditional image style transfer methods, which learn mapping between one content image and one style image, our model learn different mappings between image collections.
Such one-to-three translation task are shown in Fig.\ref{fig:extent}(a), the $\mathcal{L}_{adv}$ is redefined as
\begin{equation}
\mathcal{L}_{adv}(G,D_{ \{\text{B},\text{C},\text{D}\}  }) = \sum_{i\in \{\text{B},\text{C},\text{D}\} } (\mathbb{E}_{\chi_{i}} [\log(D_{i}(x_i))] + \mathbb{E}_{\chi_{\text{A}}}[\log(1-D_{i}(G(x_{\text{A}},z_{i})))]),
\end{equation}
where $\text{A}$ is the source domain and $\text{B}, \text{C}, \text{D}$ are target domains. In the meantime, the cycle consistency loss is modified to
\begin{equation}
\mathcal{L}_{cyc}(G) =\sum_{i\in \{\text{B},\text{C},\text{D}\} }(
\mathbb{E}_{\chi_{\text{A}}} \big[  \left \|  x_{\text{A}} - G(G(x_{\text{A}},z_{\text{i}}), z_{\text{A}})  \right \| _{1}  \big] +  
\mathbb{E}_{\chi_{\text{i}}} \big[  \left \|  x_{\text{i}} - G(G(x_{\text{i}},z_{\text{A}}), z_{\text{i}})  \right \| _{1}  \big]).
\end{equation}

\subsection{Many-to-Many Domain Translation}
As illustrated in Fig.\ref{fig:extent}(b),
the second variation shows images in multi-domain translating to each other.
In this model,
our goal is to train a single generator that can learns mappings among multiple domains and realize the mutual transformation of multiple domains.
For a four-domain transfer instance, the $\mathcal{L}_{adv}$ is redefined as
\begin{equation}
\begin{split}
\mathcal{L}_{adv}(G,D_{\{\text{A},\text{B},\text{C},\text{D}\}}) =\sum_{i,j\in \{\text{A},\text{B},\text{C},\text{D}\}}( & \mathbb{E}_{\chi_{i}} [\log(D_{i}(x_i))] + \\ &\mathbb{E}_{\chi_{j}}[\log(1-D_{i}(G(x_j,z_{i})))]),
\end{split}
\end{equation}
and the $\mathcal{L}_{cyc}$ also needs to be modified like the extended model (a).

\subsection{One-to-One Domain Translation with Multi-Modal Mapping}
To address the multi-modal image-to-image translation problem with unpaired data, we introduce the third variant as show in Fig.\ref{fig:extent}(c).
Inspired by BicycleGAN~\cite{zhu2017multimodal}, we introduce the VAE-like encoder to extract feature latent code $c$ for indicating the translation mapping.
Since there is no paired data for supervised learning of the encoder, we combine the cycle consistency with the KL-divergence to relax the constraint.
During training time, we random sample latent code from a standard Gaussian distribution $\mathcal{N}(0,I)$ to indicate the multimodality.
Then we concatenate the latent code $c$ into the domain code $z$ to indicate the final mapping.
To constraint the image content and encourage the mapping from the latent code,
we use the latent code extracted from the source image and the generated image to reconstruct the source image.

specifically,
the forward translation from domain A to domain B can be formulated as
\begin{align}
	\begin{split}
		x_{\text{B}}^{fake} = G (x_{\text{A}},z_{\text{B}},c_{\text{B}}),
	\end{split}
\end{align}
where $c_{\text{B}}$ is the latent code sampled from the prior distribution $\mathcal{N}(0,I)$.
Besides the GAN loss,
we use the latent regression to encourage the generator to utilize the latent code
\begin{equation}
	\mathcal{L}_{reg}(G,E) = \mathbb{E}_{\chi_{\text{A}},\mathcal{N}} || c_{\text{B}} - E(G(x_\text{A},z_\text{B},c_{\text{B}}),z_\text{B}) ||_1.
\end{equation}
As for the backward translation,    
the cycle reconstruction can reuse the encoder with the KL-divergence regularization.
This processes can be consisder as the cVAE~\cite{Sohn2015cVAE}
\begin{align}
\begin{split}
\mathcal{L}_{cVAE}(G,E) = \mathbb{E}_{\chi_{\text{A}},\mathcal{N}} [ &\left \| x_{\text{A}} -  G(G(x_{\text{A}},z_{\text{B}},c_{\text{B}}), z_{\text{A}},E(x_{\text{A}},z_\text{A}))  \right \| _{1} \\
&+ {KL} (E(x_{\text{A}},z_\text{A}) || \mathcal{N}(0, I) )   ],   
	\end{split}
\end{align}
Combining these two losses with the loss of base model, our model can solve the problem of the lack of diversity in unpaired image translation. Notice that we only discuss the translation of A-to-B, as the mapping of B-to-A is similar and concurrent during training time.


\section{Implementation}
\label{sec:exp}

\subsection{Network Architecture}
As in~\cite{choi2017stargan,CycleGAN2017,johnson2016perceptual,yuEccv2018},
our generator $G$ uses the ResNet\cite{he2016residual} structure with an encoder-decoder framework, 
which contains two stride-2 convolution layers for downsampling, six residual blocks and two stride-2 transposed convolution layers for upsampling.
We replace all normalization layers except upsampling layers with CBIN layers. 
For the discriminators $D$, we use two discriminators~\cite{isola2017pix2pix} to discriminate the real and fake images in different scales.
For the experiment of multi-modal SingleGAN,
the encoder model $E$ adopts the ResNet structure~\cite{zhu2017multimodal}.
We equip the encoder with CBIN, so it can also extract the latent information from different domain images.

\subsection{Training Details}
For all experiments, we train all models with Adam optimizer~\cite{kinga2015method},
setting $\beta_1 = 0.5$,  $\lambda_{cyc} = 10$,  learning rate of 0.001. 
In the extended multi-modal networks as shown in Fig.\ref{fig:extent}(c), the weights for KL-divergence and latent regression are $0.1$ and $0.5$ respectively. 
To generate higher quality results with stable training,
we replace the negative log likelihood objective by a least-squares loss~\cite{mao2017least}.

\subsection{Structure of Domain Code}
As mentioned in Sect.~\ref{sec:domain_code},
we use the one-hot vector to present the domain code $z$.
To the base model,
we use the $2$ dimensional domain code for indicating the mapping between domain $\text{A} \leftrightarrow \text{B}$.
For the one-to-many and many-to-many translation instances illustrated in Fig.~\ref{fig:extent},
the domain code is $4$ dimension and represents $4$ different domains.
In the third variant,
the $8$ dimensional latent code $c$ is also used for multimodal image translation in the specific domain that indicated by the $2$ dimensional domain code.

\begin{figure*}[!tb]
	\centering
	\includegraphics[width=0.98\linewidth]{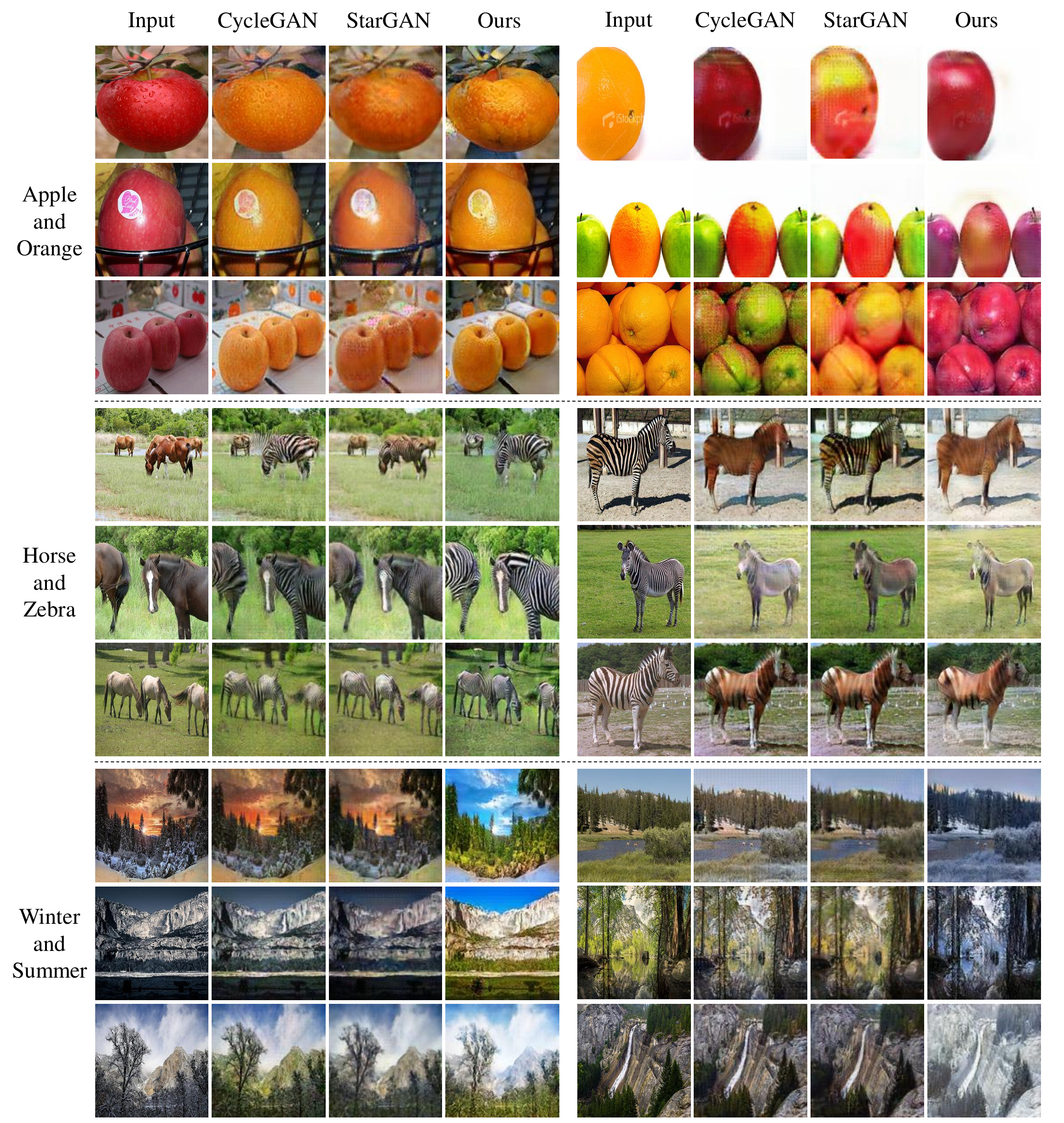}
	\caption{Visualization and comparison on three unpaired datasets. The first four columns show mapping from domain A to domain B while the next four columns show mapping from domain B to domain A.}
	\label{fig:visualization}
\end{figure*}

\section{Experiments}
\subsection{Datasets}
To evaluate the base model,
we use three unpaired datasets: Apple$\leftrightarrow$Orange, Horse$\leftrightarrow$Zebra, and Summer$\leftrightarrow$Winter~\cite{CycleGAN2017}.
As for the three extended models,
we use Photo$\leftrightarrow$Art~\cite{CycleGAN2017} for one-to-many translation,
Transient-Attributes~\cite{Laffont14} for many-to-many translation,
and Edges$\leftrightarrow$Photos~\cite{isola2017pix2pix} for one-to-one multi-model translation.
All of the images are scale to  $128 \times 128$ resolution.

\subsection{Baselines}
To compare the performance of our SingleGAN model,
we adopt the CycleGAN~\cite{CycleGAN2017} and StarGAN~\cite{choi2017stargan} as our baseline models.
CycleGAN uses cycle loss to learn the mapping between two different domains.
To achieve cycle consistency, CycleGAN requires two generators and discriminators for two different domains.
To unify multi-domain translation with single generator,
StarGAN introduces an auxiliary classifier trained on image-label pairs in its discriminator to assist the generator to learn the mapping cross multiple domains.
We compare our method with CycleGAN and StarGAN on two domains translation tasks.

\begin{table}[!b]
\begin{center}
\begin{tabular}{cccc}
\hline
  		& horse\&zebra & apple\&orange & summer\&winter \\ \hline 
image number    & 240         & 480	     &500     \\       
\hline
real image      & 0.985      & 0.978	     &0.827          \\ \hline
CycleGAN       & 0.850      & 0.935	     &0.644         \\ 
StarGAN      & 0.858 & \bf{0.970}	 &0.689     	\\
SingleGAN      & \bf{0.859} & 0.966	 &\bf{0.742}    \\ 
\hline
\end{tabular}
\end{center}
\caption{The classification accuracy for three datasets. Best results are in boldface.}
\label{tab:comparison}
\end{table}
\begin{table}[!tb]
\begin{center}
\begin{tabular}{c|c|c|c|c|c|c|c|c}
\hline
  		& \multicolumn{2}{c|}{real image}& \multicolumn{2}{c|}{CycleGAN} & \multicolumn{2}{c|}{StarGAN} & \multicolumn{2}{c}{SingleGAN} \\ \cline{2-9} 
  		&A&B 	&A&B 	&A&B &A&B \\ \hline
horse\&zebra     & 1.198 & 1.177	     &\bf{1.141} &\bf{1.133}    &1.083 &1.081   &1.112 &1.128  \\  
apple\&orange    & 1.205 & 1.499	     &1.106 &1.144   			&1.128 &1.132   &\bf{1.152} &\bf{1.164}	\\ 
summer\&winter   & 1.272 & 1.824	 	 &1.223 &1.189 	 			&1.209 &1.173 	&\bf{1.258} &\bf{1.208}  \\
\hline
\end{tabular}
\end{center}
\caption{The perceptual distance for three datasets. Best results are in boldface. Here A represents horse, apple or summer, and B represents zebra, orange or winter.}
\label{tab:distance}
\end{table}

\subsection{Base Model Comparison}
In this section, we evaluate the performance of different models.
It should be noted that both SingleGAN and StarGAN use a single generator for two domain image translation and CycleGAN uses two generators to achieve the similar mappings.

The qualitative comparison is shown in Fig.~\ref{fig:visualization}.
We can observe that all these models present pleasant results in the simple case such as the apple to orange transformation.
In the translation with complex scene,
the performance of these models are degraded especially StarGAN.
The possible reason is that the generator of StarGAN introduces the adversarial noise to fool the auxiliary classifier and fails to learn the effective mapping.
Meanwhile, we can observer that SingleGAN presents the best results in most cases.

\begin{figure*}[!b]
	\centering
	\includegraphics[width=11cm]{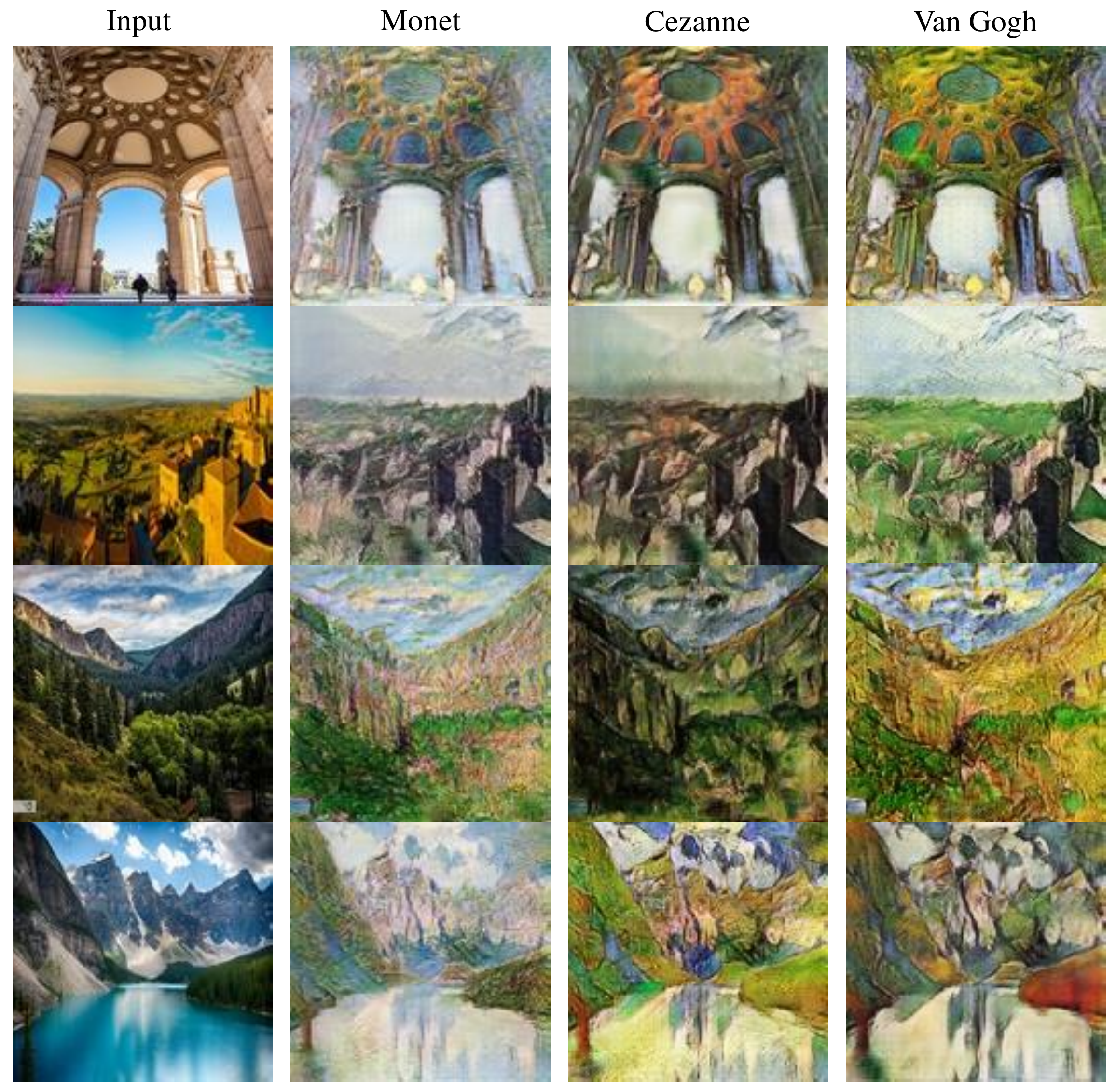}
	\caption{The one-to-many translation results of multi-style image generation. The first column is the real images and the rest of columns are the translation results that represent  different artistic styles.}
	\label{fig:style}
\end{figure*}

To judge the quality of the generated image quantitatively,
we evaluate the classification accuracy of the images generated by these three models at first.
We train three Xception~\cite{Chollet2016Xception} based binary classifiers for each image datasets.
The baseline is the classification accuracy in real images. Higher classification accuracy means that the generated images may more easy to distinguish.
Second, we compare the domain consistency between real images and generated images by computing average distance in feature space.
A similar idea is used for calculating the diversity of multi-modal generation task~\cite{zhu2017multimodal,yuEccv2018}.
we use the cosine similarity to evaluate the perceptual distance in the feature space of the VGG-16 network~\cite{Simonyan2014Very} that pre-trained in ImageNet~\cite{russakovsky2015imagenet}.
We sum across the five convolution layers preceding the pool layers.
The larger the value, the more similar between two images.
In the test stage,
we randomly sample the real image and the generated image from same domain to make up the data pair.
Then we compute the average distance between 2,000 pairs.
The baseline is computed by sampling from 2,000 pairs of real images.

\begin{figure*}[!b]
	\centering
	\includegraphics[width=\linewidth]{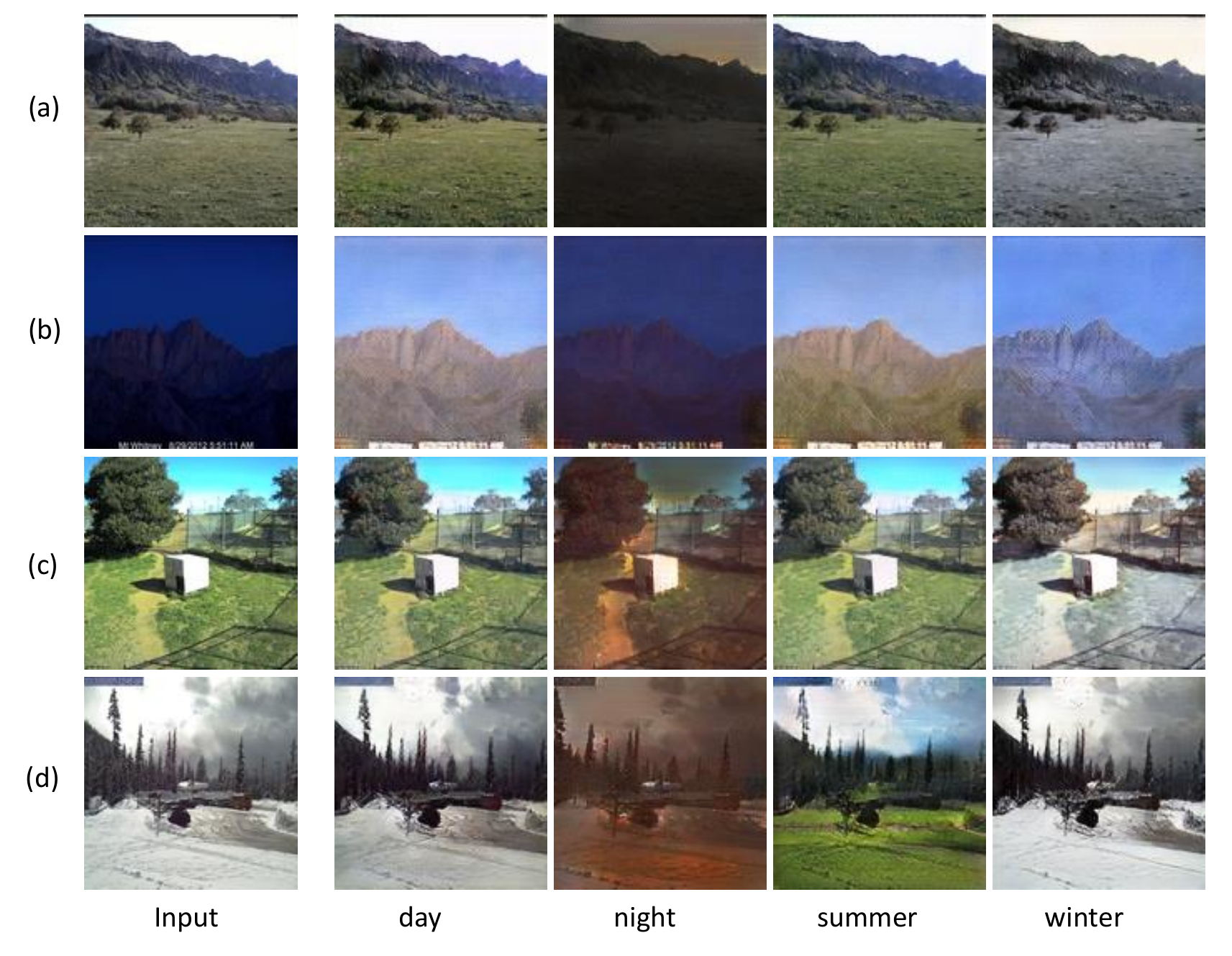}
	\caption{The many-to-many domain translation results. The first column is the input images from different domains: (a) day, (b) night, (c) summer, (d) winter. The  remaining columns are the transfer results.}
	\label{fig:face_summer}
\end{figure*}

The quantitative results are shown in Table~\ref{tab:comparison} and Table~\ref{tab:distance}.
Both SingleGAN and CycleGAN produce the quantitative results that comply with qualitative performance.
In contrast, StarGAN gets a higher classification accuracy but the poor performance in domain consistency.
It validates our conjecture that the generated image of StarGAN  may have the adversarial noise of fooling the classifier in some complex scenes.
In StarGAN, 
the discriminator learns to tell the image is real or fake without considering the classification result of the image
while the generator learns to fool the discriminator with an image that can be corrected classified by the auxiliary classifier.
So the generator may not get enough encouragement if it generates the adversarial noise to the image.
For example,
on the task of Summer$\leftrightarrow$Winter,
although the input summer image is expected to translate into winter,
the generator of StarGAN tends to 
just add a tiny adversarial noise to the input image
so that the discriminator still tell it is real
while the classifier classifies it as the winter.
As a result, the generated images will look unchanged to human but win high classification scores.
This issue does not exist in SingleGAN and CycleGAN since these models optimizes different mappings with different discriminators.
The main difference between SingleGAN and CycleGAN is the number of generators.
As shown in Fig.~\ref{fig:visualization} and Table~\ref{tab:comparison},~\ref{tab:distance},
we can observer that SingleGAN has the capacity to learn multiple mapping without performance degradation.
By sharing the generator for different domain translation,
SingleGAN can see more training data form different domains to learn the shared semantics and improve the performance of the generator.

\begin{figure*}[tb]
	\centering
	\includegraphics[width=\linewidth]{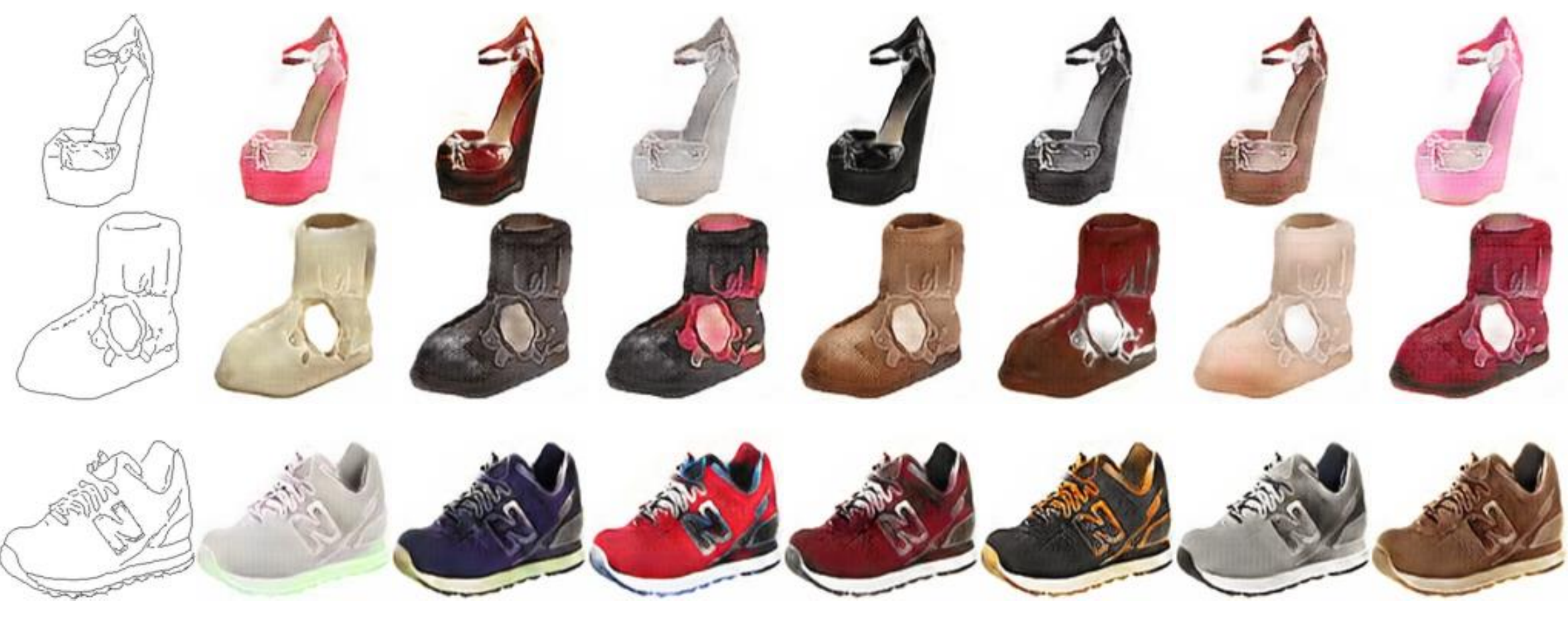}
	\caption{The multi-modal translation results. The first column shows the input and the other columns show randomly generated samples.}
	\label{fig:shoes}
\end{figure*}

\subsection{Extended Model Evaluation}
To explore the potential of SingleGAN,
we test the extended models on three different translation tasks.

For one-to-many image translation,
we perform the multi-style transfer to evaluate the model performance.
Photo$\leftrightarrow$Art~\cite{CycleGAN2017} dataset contains three artistic styles (500 images of Monet, 584 images of Cezanne and 401 images of Van Gogh) and 1000 real photos. The results are shown in Fig.~\ref{fig:style}.
We can observe that the generate images have similar artistic styles when we perform same mapping while different styles are distinguishable. 

For multi-domain translation,
we choose four outdoor scenes in Transient-Attributes dataset~\cite{Laffont14} to evaluate the model: `day',
`night', `summer' and `winter'.
It should be note that the multiple domains do not have to be independent, \eg~ the subset `day' contains summer and winter.
The training data for each domain do not need to consider other domain information.
As shown in Fig.~\ref{fig:face_summer}, SingleGAN is competent at the transformation from all domains, though the dataset has incomplete labels.

The final experiment is to verify the multi-modal performance of SingleGAN after introducing the attribute latent code.
The dataset adopted is edge2shoes~\cite{zhu2016generative}.
Please Note that this experiment is performed under the settings of unpaired data.
The experimental result in Fig.~\ref{fig:shoes} shows that SingleGAN has the ability to learn multimodal mapping under the unsupervised learning.

\begin{figure*}[!t]
	\centering
	\includegraphics[width=\linewidth]{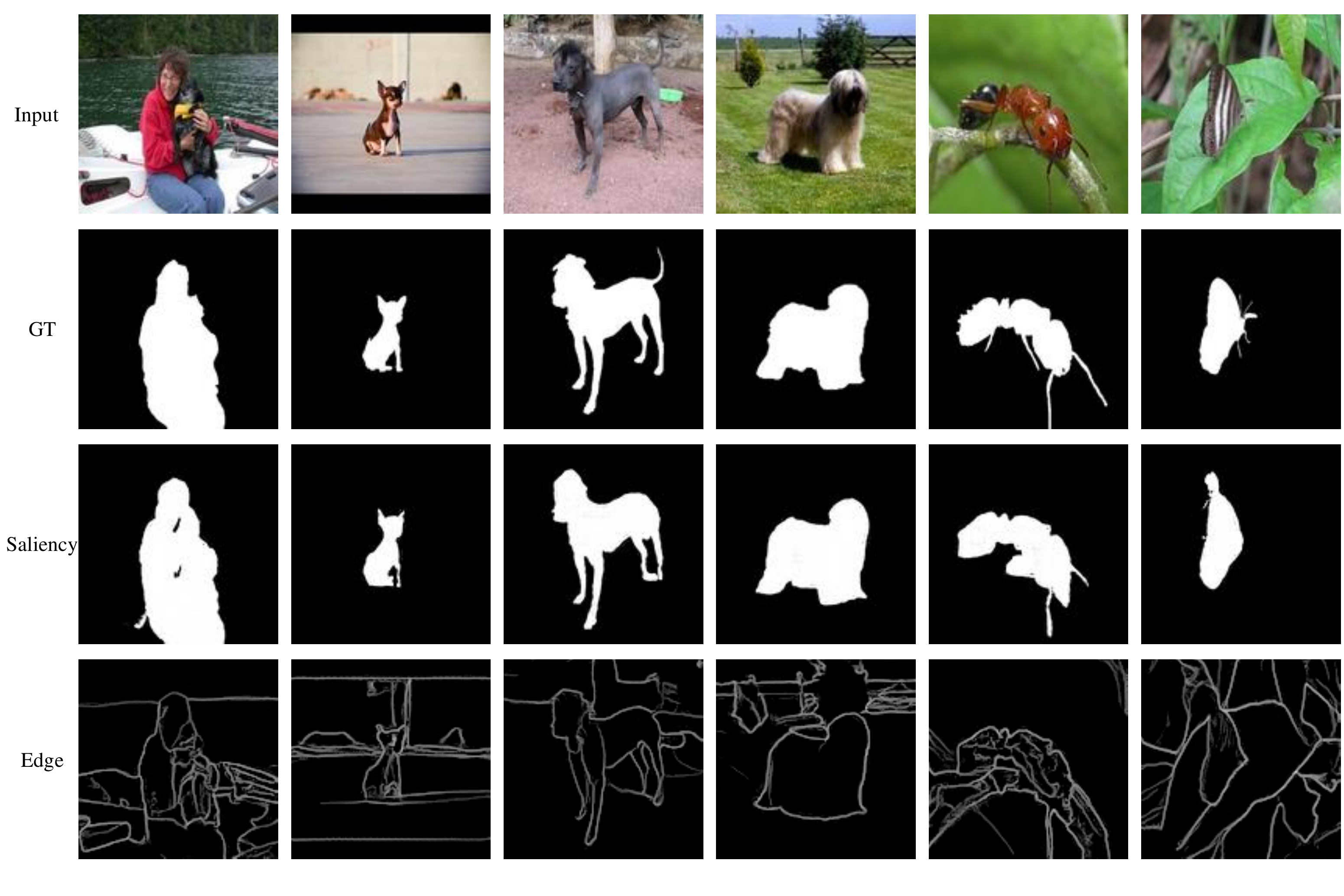}
	\caption{Saliency and edge detection results of SingleGAN under paired data setting.}
	\label{fig:Saliency}
\end{figure*}

\subsection{Translation under Paired Data Setting}
Although the above experiments have the unpaired data assumption,
SingleGAN can also perform multi-domain image translation with paired data by replacing the cycle consistency loss $\mathcal{L}_{cyc}$ with $\ell_1$ reconstruction loss $\mathcal{L}_{rec}$.

Here we use the salient object dataset DUTS-TR~\cite{zhao2015saliency} and BSDS500 edge dataset~\cite{arbelaez2011contour} to perform one-to-many image translation.
Specify the real image as domain A, salient images as domain B and edge images as domain C.
Then the $\mathcal{L}_{rec}$ can define as 
\begin{equation}
\begin{split}
\mathcal{L}_{rec}(G) =&  
\mathbb{E}_{\chi_{\text{A}},\chi_{\text{B}} } \big[  \left \| x_{\text{B}} -  G(x_{\text{A}},z_{\text{B}})  \right \| _{1}  \big] + \\ 
&\mathbb{E}_{\chi_{\text{A}},\chi_{\text{C}} } \big[  \left \| x_{\text{C}} - G(x_{\text{A}},z_{\text{C}}) \right \| _{1}  \big].
\end{split}
\end{equation}
The results in Fig.~\ref{fig:Saliency} demonstrate the effectiveness of SingleGAN. 

\subsection{Limitations and Discussion}
Although SingleGAN can achieve multi-domain image translations,
multiple adversarial learning needs to be done simultaneously.
This constraint makes SingleGAN only be able to learn limited domain translation at a time since our storage is limited.
So it is valuable to explore the transfer learning for the existing models.
Besides,
the capacity of the network to learning different mappings is also an important problem.
We also observe that integrate suitable tasks for one single model may improve the performance of the generator.
But what kind of tasks can promote each other remains to be explored in the future work.
Nonetheless, we think the method proposed in this paper is valuable for exploring the multi-domain generation works. 

\section{Conclusion}
\label{sec:conc}
In this paper we introduce a single generator based model, SingleGAN, for learning multi-mapping image-to-image translation.
By introducing multiple adversarial learning for the generator,
SingleGAN is able to learn a variety of mappings effectively and efficiently.
Contrastive experimental results show quantitatively and qualitatively that our approach is effective in many image translation tasks.
Furthermore, to improve the versatility and generality of the model,
we present three variants of SingleGAN for different tasks:
one-to-many domain transfer,
many-to-many domain transfer and one-to-one domain transfer with varying attributes.
The experiment results demonstrate these variants improve the corresponding translation effectively.

\subsubsection{Acknowledgments.}
{This work was supported in part by the Project of National Engineering Laboratory for Video Technology - Shenzhen Division,
National Natural Science Foundation of China and Guangdong Province Scientific Research on Big Data (No.U1611461),
Shenzhen Municipal Science and Technology Program under Grant JCYJ20170818141146428,
and Shenzhen Key Laboratory for Intelligent Multimedia and Virtual Reality (No.ZDSYS201703031405467).
}

\bibliographystyle{splncs}


\end{document}